\pdfoutput=1

\documentclass[11pt]{article}

\usepackage{ACL2023}

\usepackage{times}
\usepackage{latexsym}

\usepackage[T1]{fontenc}

\usepackage[utf8]{inputenc}

\usepackage{microtype}

\usepackage{inconsolata}

\usepackage{nicematrix}
\usepackage{caption}
\usepackage{subcaption}
\usepackage{placeins}
\usepackage{enumitem}

\title{Retrieval-Based Transformer for Table Augmentation}

\author{Michael Glass$^1$,
Xueqing Wu$^2$,
Ankita Rajaram Naik$^1$,
Gaetano Rossiello$^1$,
Alfio Gliozzo$^1$ \\
$^1$IBM Research AI, Yorktown Heights, NY, USA \\
$^2$University of Illinois Urbana-Champaign
}

\newcommand{\retwog}{$\text{Re}^2\text{G}$}

\newcommand{\lcandidate}{$\langle$}
\newcommand{\rcandidate}{$\rangle$}
\newcommand{\webtables}{WebTables}

\newcommand{\xueqing}[1]{}
\newcommand{\ankita}[1]{}

\begin{document}

\maketitle

\begin{abstract}

Data preparation, also called data wrangling, is considered one of the most expensive and time-consuming steps when performing analytics or building machine learning models.
Preparing data typically involves collecting and merging data from complex heterogeneous, and often large-scale data sources, such as data lakes.
In this paper, we introduce a novel approach toward automatic data wrangling in an attempt to alleviate the effort of end-users, e.g. data analysts, in structuring dynamic views from data lakes in the form of tabular data.
We aim to address \textit{table augmentation} tasks, including row/column population and data imputation.
Given a corpus of tables, we propose a retrieval augmented self-trained transformer model. 
Our self-learning strategy consists in randomly ablating tables from the corpus and training the retrieval-based model to reconstruct the original values or headers given the partial tables as input. 
We adopt this strategy to first train the dense neural retrieval model encoding table-parts to vectors, and then the end-to-end model trained to perform table augmentation tasks.
We test on EntiTables, the standard benchmark for table augmentation, as well as introduce a new benchmark to advance further research: WebTables.
Our model consistently and substantially outperforms both supervised statistical methods and the current state-of-the-art transformer-based models.

\end{abstract}

\section{Introduction}
The way organizations store and manage data is rapidly evolving from using strict transactional databases to data lakes that typically consist of large collections of heterogeneous data formats, such as tabular data, spreadsheets, and NoSQL databases.
The absence of a unified schema in data lakes does not allow the usage of declarative query languages, e.g. SQL, making the process of data preparation\footnote{Also referred as data wrangling or data munging.} dramatically expensive~\cite{terrizzano2015data}.

Data preparation involves several phases, such as data discovery, structuring, cleansing, enrichment and validation, with the purpose of producing views commonly organized in a tabular format used to create reports~\cite{DBLP:journals/tbd/KoehlerABCMKFKL21} or to gather feature sets to build machine learning models~\cite{DBLP:journals/kbs/HeZC21}.
The schemaless nature of data lakes makes data discovery and structuring even more challenging since the tasks of joinability and unionability among tables become non-deterministic~\cite{DBLP:conf/icde/FernandezAKYMS18,DBLP:conf/sigmod/ZhuDNM19,DBLP:conf/icde/BogatuFP020}.

In this work, we propose a novel end-to-end solution based on a retrieval augmented transformer architecture with the aim to support end-users, such as data analysts, in the process of constructing dynamic views from data lakes.
To this end, we address three table augmentation tasks~\cite{entitables,zhang2019auto}: automatic row and column population and cell filling (or data imputation).

\begin{figure*}[tbh]
  \centering
   \includegraphics[width=0.8\linewidth]{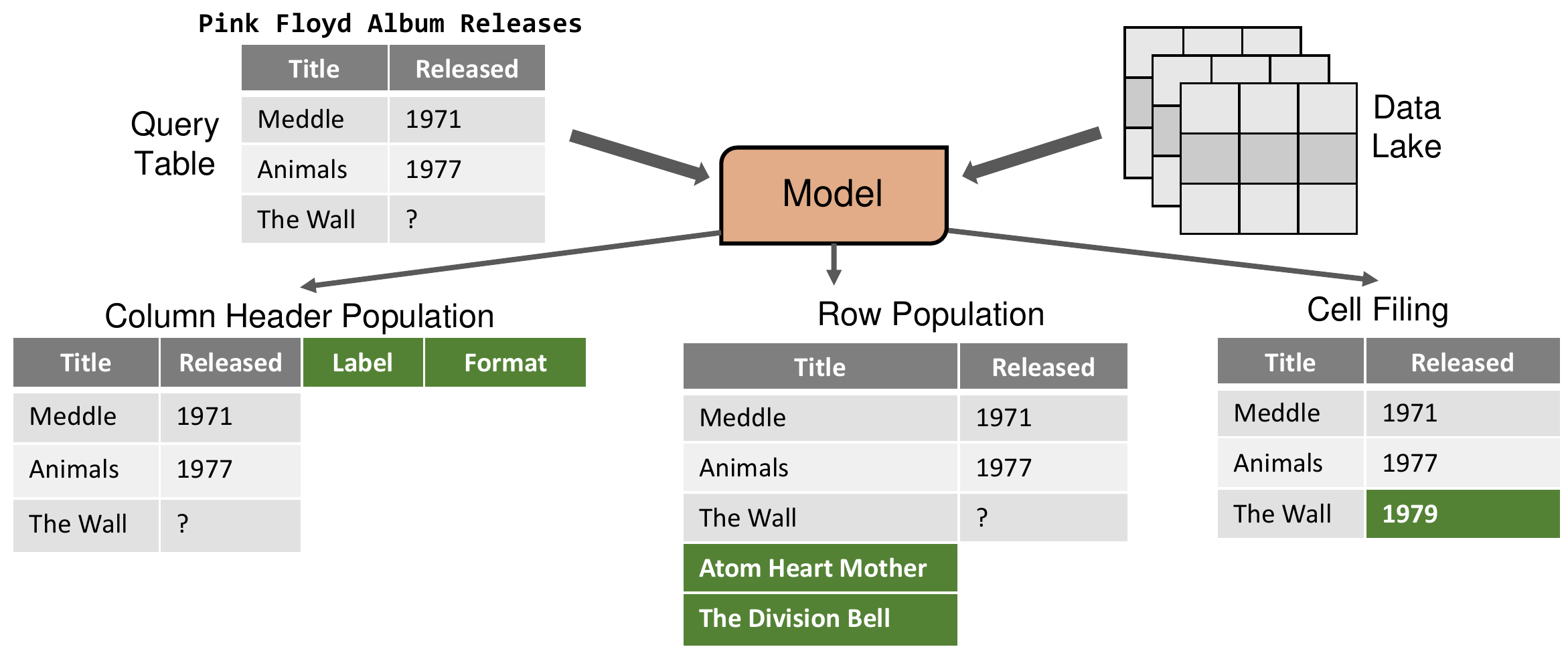}
   \caption{Given a partially completed table as a query (i.e. a few album releases from the Pink Floyd discography), the three table augmentation tasks consist of retrieving from the data lake: 1) a list of possible next column headers, such as the ``Label" or ``Format", 2) the missing value ``1979" for the release date of the row ``The Wall", 3) a list of other album releases as possible next rows, such as ``Atom Heart Mother" and ``The Division Bell".}
   \label{fig.task}
\end{figure*}

Figure \ref{fig.task} illustrates the three core tasks in table augmentation. All tasks proceed from a query or seed table. In the case of self-supervised training, this seed table is formed by ablating rows, columns or cell values from an existing table in the data lake.
The task of column header population, also simply called column population, is to extend the table with additional possible column names or headers. This is a way of suggesting additional data that could be joined into this table.
In the task of cell filling there is a specific unknown cell, for which the model predicts a specific value.
The task of row population 
is only populating the \textit{key column} for a row. This is the column that contains the primary entity that the remainder of the row contains data for, sometimes referred to as a row header. Typically this is the first column in a table.

Approaches to table augmentation can be purely parametric~\cite{tabbie,turl}, in which case the data lake is used to train the parameters of the model, but not used during inference. In this setting, the table augmentation model must draw the possible augmentations for rows, columns and cells from its trained parameters. 
Alternatively, with retrieval-based models~\cite{rag,kgi_emnlp,re2g}, the data lake can also be used at inference to provide evidence for proposed augmentations. This has two key advantages: 1) the model need not memorize the data lake -- or even a significant fraction of it, and 2) the model can provide justification for its predicted augmentations in the form of a provenance table or tables.

The key contributions of this paper are:
(1) We introduce the first end-to-end, retrieval-based model for table augmentation. Our Retrieval Augmented Table Augmentation (RATA) model uses a biencoder retrieval model for neural indexing and searching tables from data lake, and a reader transformer to identify augmentations from retrieved tables.
(2) Our model establishes a new state-of-the-art across all three tasks in table augmentation, while also providing additional value with its provenance information.
(3) We create and release a new dataset for table augmentation, expanding the scope of evaluation beyond Wikipedia. This dataset, based on \citet{webtables}, is also larger and more diverse than the standard Wikipedia-based dataset~\cite{entitables}.


\section{Related Work}\label{sec.relatedwork}
\paragraph{Table augmentation} can be divided into three sub-tasks: row population, column population, and cell filling.
For row and column population, \citet{entitables} identifies and ranks candidate values from both the table corpus and knowledge base. Table2Vec \citep{table2vec} trains header and entity embeddings from a table corpus in a skip-gram manner and uses the embeddings for the task. Although TaBERT~\cite{tabert} was developed as a foundational model primarily for question answering, its embeddings have also been applied for row and column population. Recent work formulates the task as multi-label classification and fine-tunes large-scale pre-trained models such as TABBIE~\citep{tabbie} and TURL~\citep{turl}.

TABBIE consists of three transformers for converting cells, columns and rows to vector representations. A corrupt cell detection task is the pretraining task used to learn these embeddings on the table corpus. 
To fine-tune a trained TABBIE model for the column header population task, a concatenated [CLSCOL] embedding of the columns is passed through a single linear and softmax layer and trained with a multi-label classification objective. Similarly, for the row population task a multi-class classification is carried out on the first column's [CLSCOL] representation.

\begin{figure*}[th!]
\centering
  \begin{subfigure}[b]{0.4\linewidth}
    \centering
    \includegraphics[width=\linewidth]{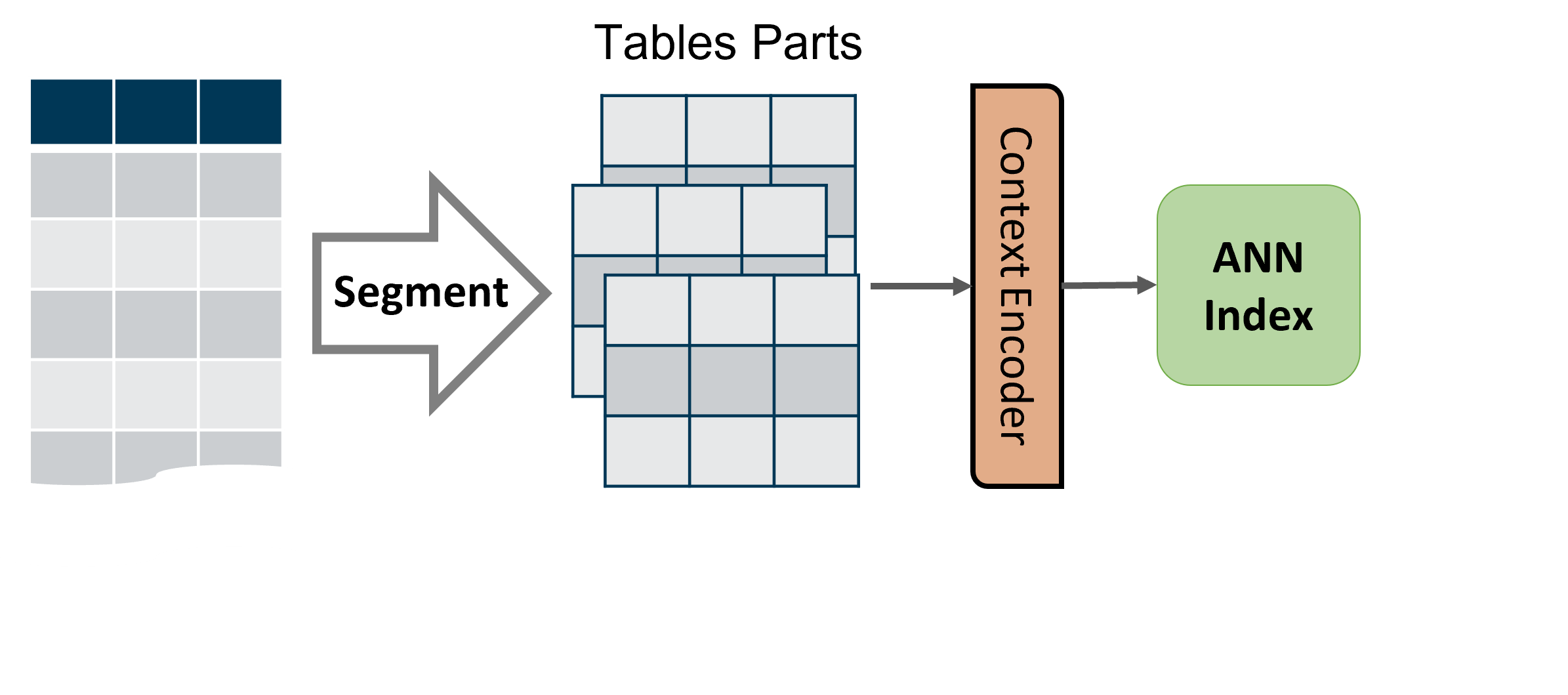}
    \caption{Building the table-part index.}\label{fig.tablecorpus}
  \end{subfigure}
  \begin{subfigure}[b]{0.5\linewidth}
    \centering
    \includegraphics[width=\linewidth]{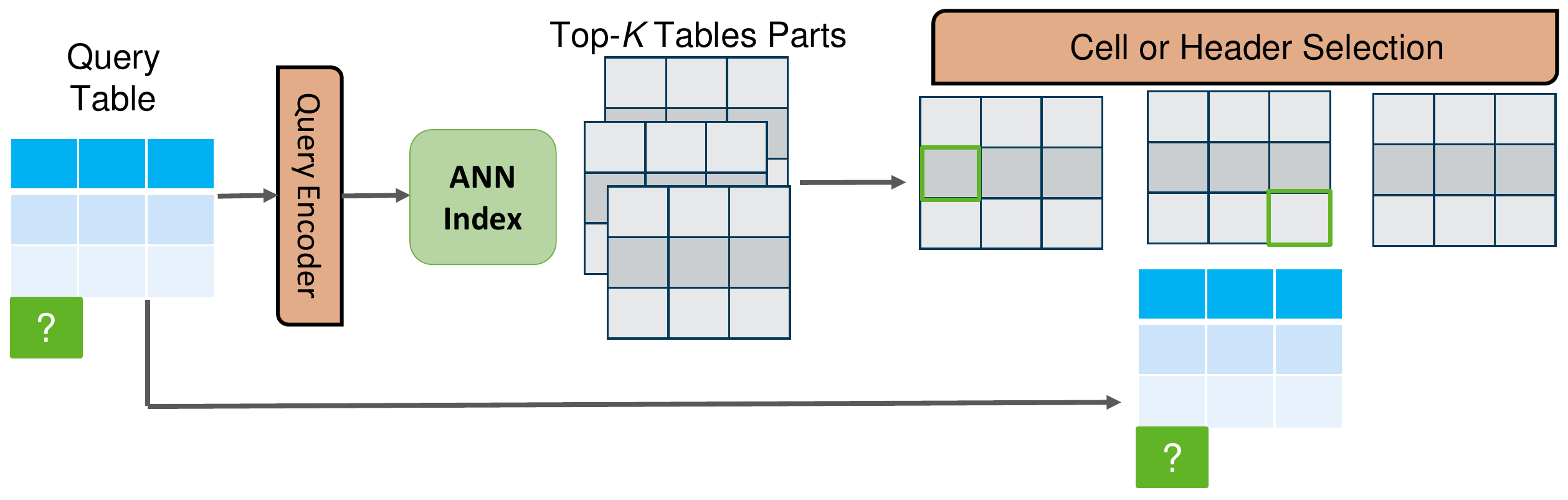}
    \caption{Architecture of RATA.}\label{fig.architecture}
  \end{subfigure}
  \caption{Index building and inference system overviews}
\end{figure*}

For cell filling, InfoGather \citep{infogather} retrieves tables from the table corpus and selects values from retrieved tables. \citet{zhang2019auto} extends the system to retrieve from both the table corpus and knowledge base. 
Their system that uses only the table corpus as the source is called TMatch, which we compare to in Section \ref{sec:results}.
\citet{hybrid_imputation} combines predictions both from table retrieval and from a machine learning-based value imputation system.
\citet{turl} directly applies pre-trained TURL model to the task since cell filling is similar with its pre-training objective.
Cell filling is also related to the task of value imputation, i.e., to provide an assumed value when the actual value is unknown, usually using machine learning methods \citep{datawig}. 
In addition to augmenting individual entities, column headers or cells, some other work aims to join tables over entire rows or columns with retrieved tables \citep{DBLP:conf/sigmod/SarmaFGHLWXY12,DBLP:conf/kdd/BhagavatulaND13,DBLP:journals/ws/LehmbergRRMPB15}.

\paragraph{Retrieval-augmented models} have been successfully applied to many tasks. For open-domain question answering (ODQA), DPR learns dense representation to retrieve evidence and trains a separate reader to select answer from retrieved evidence \citep{dpr}. RAG uses a generator to generate outputs conditioned on retrieved evidence and jointly trains DPR with a generator on the downstream task \citep{rag}. RAG is shown to achieve good performance on knowledge-intensive NLP tasks such as ODQA, fact verification, slot filling, etc \citep{rag,kilt}. \retwog{} further introduces a reranker to boost performance \citep{re2g}. Retrieval-augmented models are also shown to be effective on zero-shot slot filling \citep{kgi_emnlp}, and multilingual keyphrase generation \citep{gao-etal-2022-retrieval}.
Similar models have also been applied to table-related tasks such as open-domain table question answering \citep{nqtable}. In our work, we apply the architecture to table augmentation.

\section{Approach}\label{sec.approach}

While the row, column, and cell predictions of purely parametric table augmentation methods may be useful on their own, they can be much more effective for a human-in-the-loop use case if they are supported by provenance.
A user of a data preparation application may be unwilling to simply accept the prediction of a model, but when paired with evidence from the data lake, that prediction can be better assessed.
Furthermore, the retrieval model itself may be useful for exploration and general search in a data lake. In this view, table augmentation can be seen as self-supervised pretraining for table retrieval.

Fortunately, there is now considerable work on \textit{retrieval augmented} transformer models \citep{re2g,rag}. These models augment the parametric knowledge of the transformer, with non-parametric knowledge in the form of an indexed corpus. To do so, they 
use a neural retrieval model based on DPR (Dense Passage Retrieval) ~\cite{dpr} that is trained end-to-end to assist in generation.

We build on this line of research to introduce a general model for all table augmentation tasks: row population, column header population and cell filling. Our model, Retrieval Augmented Table Augmentation (RATA), comprises of an index of tables, a retrieval component, and a reader or selection component.
The table index is built from the tables in the training set, which are first decomposed into table-parts, then transformed into sequences for use with standard retrieval approaches.
The retrieval component is a biencoder architecture similar to DPR \cite{dpr}, but trained without ground truth on correct provenance.  We call this \textit{Dense Table Retrieval} or DTR.
The reader component is an extractive approach.  An extractive rather than generative approach ensures that the model's predictions are always grounded in actual data, rather than speculative guesses. The extractive approach is also a more natural fit for row and column population tasks, where there is no required order to the answers. Finally, the extractive approach permits an initial training phase for the retrieval component where the \textit{answer-bearing} tables are considered as a bag of positives.

Figure \ref{fig.task} illustrates the tasks of table augmentation by example. Formally, the input $I$ is a table with $r$ rows and $c$ columns comprising a caption $\mathcal{C}$, headers $\mathbf{H}$, and matrix of cell values, $\mathbf{V}$. One of the columns, usually the first, is indicated as the key column $key$.
\begin{align*}
I & = \langle \mathcal{C}, \mathbf{H}, \mathbf{V}, key \rangle, 1 \leq key \leq c \\
\mathbf{H} & = [h_1, h_2, ..., h_{c}] \\
\mathbf{V} & = \begin{bmatrix}
v_{1,1}, v_{1,2}, ..., v_{1,c} \\
... \\
v_{r,1}, v_{r,2}, ..., v_{r,c} \\
\end{bmatrix}
\end{align*}

The input table is ablated in a task specific way to produce a query table and gold answers, $\langle Q, \mathbf{G} \rangle$, described as follows:
\begin{align*}
Q_{rp} & = \langle \mathcal{C}, \mathbf{H}, \mathbf{V_{ 1..n_{seed}}}, key \rangle \\
\mathbf{G}_{rp} & = \{ \mathbf{V}_{i,key} : i > n_{seed} \} \\
Q_{cp} & = \langle \mathcal{C}, \mathbf{H_{1..n_{seed}}}, \mathbf{V_{.., 1..n_{seed}}}, key \rangle \\
\mathbf{G}_{cp} & = \{ \mathbf{H}_i : i > n_{seed} \} \\
Q_{cf} & = \langle \mathcal{C}, \mathbf{H}, \mathbf{V \setminus v_{i,j}}, key \rangle \\
\mathbf{G}_{cf} & = \{ v_{i,j} \} 
\end{align*}
where \textit{rp}, \textit{cp} and \textit{cf} refer to the row population, column header population and cell filling tasks, respectively. 

\subsection{End-to-End Model}\label{sec:e2emodel}

Figure \ref{fig.tablecorpus} shows how tables in a data lake are first indexed to provide a non-parametric knowledge store.  Each table is first split into chunks of up to three rows plus the header, which we refer to as \textit{table-parts}. We form sequence representations of these table-parts following work in other transformer-based approaches to tables~\cite{rci}.
The table-part sequence representations ($S^{t}$) are formed from the row sequence representations ($S^{r}_{i}$) and the table caption:
\begin{align*}
S^{r}_{i} & = \bigoplus_{j=1}^{c} h_j \oplus \text{`:'} \oplus\ v_{i,j} \oplus \text{`*'} \\
S^{t} & = \mathcal{C} \oplus \text{[SEP]} \oplus \bigoplus_{i=start}^{end} S^{r}_{i} \oplus \text{`|'}
\end{align*}
Here $\oplus$ indicates concatenation and the strings `:', `*', and `|' delimit the header, cell value contents, and each row respectively.  Any distinctive tokens can work as delimiters since the transformer will learn an appropriate embedding representation.

These sequences are then projected to vectors using the context encoder by taking the [CLS]. We index the dense representations for all table-parts in the data lake using FAISS~\cite{faiss} with Hierarchical Navigable Small World~\cite{hnsw}. 

Figure \ref{fig.architecture} shows the architecture of our approach, Retrieval Augmented Table Augmentation (RATA).
The input query is encoded to a vector for retrieving related table-parts from the indexed data lake. 
Similar to table-part representation, we form sequence representation for the query, use a query encoder to encode it, and take the [CLS] vector as query representation. Both the context encoder and the query encoder use the BERT$_{\text{BASE}}$ architecture. We use unnormalized dot product to score a pair of query $q$ and table-part $d$. Top-k table-parts with highest scores will be retrieved.
\begin{align*}
score(q, d) & = \text{BERT}_{qe}(q)_{[CLS]} \cdot \text{BERT}_{ce}(d)_{[CLS]} 
\end{align*}
After the top-k most relevant table-parts are retrieved, the reader component selects the most likely augmentations for the query table. In the case of column population, the candidate augmentations are all headers from retrieved table-parts; for cell filling it is all cells; and for row population it is only those cell values that are entities.

The sequence representation of the query table is paired with each table-part representation, using the standard [CLS] and [SEP] token to demarcate the bounds of each sequence. 
In the table-part representation, the candidates are marked by special begin and end tokens: `\lcandidate{}' and `\rcandidate{}'.
This combined sequence is then the input to a transformer encoder (initialized from BERT$_{\text{LARGE}}$ \cite{bert}). For each pair of candidate answer marks (`\lcandidate{}' and `\rcandidate{}'), the final token embeddings  are concatenated to produce a single vector. Then a linear layer is applied to predict the likelihood that the candidate is a correct answer to the query. 
\begin{align*}
\alpha & = [i : t_{i} = ``\langle" ] \\
\omega & = [i : t_{i} = ``\rangle" ] \\
ans_n & = t_{\alpha_n + 1}, t_{\alpha_n + 2}, ..., t_{\omega_n - 1} \\
C & = \begin{bmatrix}
E_{\alpha_0} \oplus E_{\omega_0} \\
E_{\alpha_1} \oplus E_{\omega_1} \\
E_{\alpha_2} \oplus E_{\omega_2} \\
...
\end{bmatrix} \\
\rho & = softmax(C \cdot \mathbf{w_{candidate}})
\end{align*}

Formally, the input is a sequence of tokens $T = [t_0, t_1, ...]$. The transformer encoder produces a sequence of embeddings $BERT_{reader}(T) = E = [e_0, e_1, ...]$. The candidate representation vectors, $C$, are then multiplied by the learned parameter vector $\mathbf{w_{candidate}}$ and a softmax is applied to produce the reader scores, $\rho$, for the retrieved table-part.

Note that the likelihood for a given answer occurrence $ans_n$ is $\rho_n$. The candidate likelihood vectors for each of the top-k retrieved table-parts, $\rho^1, \rho^2, ..., \rho^k$, are then combined with the softmax normalized retrieval scores, $\mathbf{r} = [r_1, r_2, ..., r_k]$, to provide a probability distribution over all candidates in all retrieved table-parts.
Since these scores are for each occurrence of a candidate string, we aggregate over each distinct normalized candidate string by summing the likelihoods for all occurrences. This produces the final score, $s(a)$ for each answer string $a$. The loss is the negative log-likelihood of all gold answer strings, $\mathbf{G}$. Because of this formulation, during training any instance with no correct candidates in any retrieved table-part is skipped.
\begin{align*}
\mathbf{p}^j & = softmax(\mathbf{r})_j \cdot \rho^j \\
s(a) & = \sum_{j=1}^{k} \  \sum_{n : ans^j_n = a} \mathbf{p}^j_n \\
loss & = -\sum_{a \in \mathbf{G}} log \left( s(a) \right)
\end{align*}
We use answer normalization to determine if a candidate matches a gold answer, as described in Appendix \ref{apx.dataset_details}. For row population and cell filling in EntiTables, the cell values are already linked to entities so normalization is not necessary. 

For RATA training, we iterate through the tables in the training set. To construct input query from a table, we ablate either all rows after the first $n_{seed}$ (row population), or all columns after the first $n_{seed}$ (column population), or a particular cell (cell filling). We ensure that table-parts from the query table are not retrieved by filtering the retrieved results. Like most previous approaches to end-to-end training of neural retrieval, we train only the query encoder in the end-to-end training phase. This avoids expensive re-indexing of the entire data lake either each time the context encoder is updated, or periodically as in ANCE~\cite{ance}.

\subsection{Retrieval Training}\label{sec:retrieval}


While it is possible in theory to train neural retrieval entirely through impact in the end-to-end table augmentation tasks, a good initialization is important for learning.  Without an initial effective retrieval model, there is no answer-bearing evidence to train the reader model, and therefore a high fraction of training examples will be skipped~\cite{orqa}. 

One possible approach is to use a pretraining task for retrieval, such as the Inverse Cloze Task~\cite{orqa} or a retrieval-based masked language model~\cite{realm}. In the table augmentation task, there is the option of training with answer-bearing evidence as positives.  Since the reader is purely extractive, any evidence that does not contain a correct augmentation string is necessarily a negative. However, not every table-part that contains an answer is a positive.  We use a multiple instance learning setup for the positives: we train under the assumption that at least one of the table-parts containing a correct answer is a positive.

To gather the training data for retrieval we build an initial keyword index using Anserini\footnote{\url{https://github.com/castorini/anserini}}. We use BM25~\cite{bm25} to retrieve potentially relevant table-parts for each table query.

From each training table we construct a query for row population, column population or cell filling. Since these queries are constructed from ablated tables, we know a (potentially incomplete) set of correct augmentations or answers. Note that there may be other equally correct augmentations. But since this is a self-supervised task, we consider only the headers or cell values that actually occurred in the table to be correct.

Formally, the query constructed from a training table is a pair of the ablated table, $Q$ and the set of gold answers $\mathbf{G}$. The set of table-parts retrieved by the initial retrieval method, for example BM25, is given as $\mathbf{R}$. A retrieved table-part is in the positive set, $\mathbf{R}^+$, if it contains any gold answer, otherwise it is a hard negative, $\mathbf{R}^-$.
\begin{align*}
\mathbf{R}^+ & = \{ d : d \in \mathbf{R} \wedge \exists a \in \mathbf{G}, a \in d \} \\
\mathbf{R}^- & = \mathbf{R} - \mathbf{R}^+
\end{align*}

Following \citet{dpr}, we use batch negatives along with the retrieved ``hard negatives''. The batch $B = [\langle q_1, \mathbf{R}_1 \rangle, \langle q_2, \mathbf{R}_2 \rangle, ..., \langle q_{bz}, \mathbf{R}_{bz} \rangle]$ is processed to produce vectors for all queries and retrieved table-parts. All query vectors are multiplied with all table-part vectors to produce scores between all pairs. A softmax is applied per-query to give the normalized scores. Finally, the loss is the negative log-likelihood for the positive scores.
\begin{align*}
\mathcal{R} & = \bigcup_{i = 1}^{bz} \mathbf{R}_i \\
\rho_i & = softmax([score(q_i, d) : d \in \mathcal{R}]) \\
loss & = - \sum_{i = 1}^{bz} log \left( \sum_{d \in \mathbf{R}^+_i} \rho_{i,d} \right)
\end{align*}

Note that since we are summing over the probability of all table-parts in the positive set, $\mathbf{R}^+$, it is not necessary for \textit{all} answer-bearing retrieved table-parts to be high scoring. 
Instead, it follows the multiple instance learning framework. All instances marked negative are negative, while at least one instance in the positive set is positive.

\section{\webtables{} Dataset}\label{sec:webtables}

Prior work on table augmentation has focused on tables derived from Wikipedia \cite{entitables, tabbie, turl, zhang2019auto, zhang2019table2vec}. In order to better assess the proposed methods and provide the research community with a new benchmark, we introduce a new dataset for table augmentation: \webtables.

We construct this dataset using the tables crawled and extracted by \citet{webtables}. We start from the English relational tables of WDC Web Table Corpus 2015. We further filter the dataset to remove the most common types of noisy tables: calendars formatted as tables, lists of forum posts and torrent links, tables with less than four rows or columns, and tables that format large blocks of text. Because previous work on table augmentation focused so heavily on Wikipedia tables, we exclude from this dataset any tables crawled from any ``wikipedia'' domain.  We also deduplicate the corpus, ensuring that there are no two tables with the same content in their cells.

Following filtering and deduplication we sample 10 thousand tables each for the development and test sets and one million tables for training. However, in our experiments we use only 300 thousand training examples to limit the computational cost.

To parallel the setting of EntiTables we use the ``key column'' identified by \citet{webtables} as the target column for row population and we consider entities to be those strings that occur at least three times in the key column for any table in the train set.

\section{Experiments}\label{sec:experiments}

We experiment on two datasets of tables across three tasks. Table \ref{tab:datasets} gives statistics on these datasets.

\textbf{EntiTables} \citep{entitables} contains 1.6M tables collected from Wikipedia where entity mentions are normalized into its name in DBPedia.
For row and column population, we use the development and test sets released by \citet{entitables} each containing 1,000 randomly sampled queries.
For cell filling, we use the test set released by \citet{zhang2019auto}. The test set contains 1,000 queries uniformly sampled from four main column data types: entity, quantity, string, and datetime. Though \citet{zhang2019auto} use human annotations as gold labels, we notice that the human annotations are of low quality, so we use the original values in the table cells as gold labels.

\textbf{\webtables{}} is based on \citet{webtables} -- 154M relational tables extracted from HTML tables in Common Crawl. We process the corpus as described in Section \ref{sec:webtables}. 
For column population we use the original development and test sets of 10,000 tables each. While for row population we necessarily exclude any tables without any entities in the key column after the first $n_{seed}$ rows. 
For cell filling, we use heuristic rules to classify cell values into three types: quantity, string and datetime. Then, we sample 3,000 queries uniformly from the three types as test set and sample another 3,000 queries as development set.

\begin{table}[!htb]
\centering
\begin{NiceTabular}{ll|rrr}
\hline
\textbf{Dataset} & \textbf{Task} & \textbf{Train} & \textbf{Dev} & \textbf{Test} \\
\hline
EntiTables & \small{row pop.} & 187k & 1k & 1k \\
EntiTables & \small{column pop.} & 602k & 937 & 950 \\
EntiTables & \small{cell filling} & 100k & - & 972 \\
\hline
\webtables{} & \small{row pop.} & 563k & 6.6k & 6.8k \\
\webtables{} & \small{column pop.} & 1M & 10k & 10k \\
\webtables{} & \small{cell filling} & 1M & 3k & 3k \\
\hline
\end{NiceTabular}
\caption{Dataset statistics.}
\label{tab:datasets}
\end{table}


We compare our method with two deep learning-based baselines, TABBIE \citep{tabbie} and BART \citep{bart}. Both TABBIE and BART have no retrieval component involved.  

TABBIE, described in Section \ref{sec.relatedwork}, uses three transformers: one for cell values, one for rows, and one for columns. It produces vector embeddings for each cell and each row and column of a table.
We follow \citet{tabbie} for the row and column population and base our experiments on the partial released code and pretrained model\footnote{https://github.com/SFIG611/tabbie}.
To apply TABBIE to cell filling, we formulate it as classification on the concatenation of the row and column embedding vectors, similar to row and column population. The classification vocabulary is collected from the training corpus: all cell values that occur at least ten times. 
We also report the published results for TABBIE on the EntiTables dataset, although we were unable to reproduce these results for row population.

BART is a sequence-to-sequence model that takes the linearized table as the source text and generates the row entities, cell headers, or cell value as the target text.  We use a beam search in decoding (beam size = 35) to produce a ranked list of predictions.  We use the FAIRSEQ toolkit~\cite{fairseq} for these experiments.  For RAG we use the implementation in Hugging Face transformers~\cite{huggingface}. For both BART and RAG, the sequence representation of the query tables is the same as in RATA.   

On the EntiTables dataset, we also compare against probabilistic methods that first retrieve tables from the table corpus and next select values for table augmentation. We compare against the published results of \citet{entitables} for row and column population, and against TMatch~\cite{zhang2019auto} for cell filling.

For evaluation, we report Mean Reciprocal Rank (MRR) and Normalized Discounted Cumulative Gain over the top ten outputs (NDCG@10) for the final prediction performance of row population, column population, and cell filling. To evaluate the performance of DTR retrieval, we also report answer-bearing MRR, where a retrieved table-part is considered correct if it contains one of the correct answers. To determine the significance of these results we use a 95\% confidence interval on the t-distribution. We also applied a sampling permutation test, but this did not change any conclusions regarding significance.








\section{Results}\label{sec:results}



Table \ref{tab:row_pop} contains our results for the row population task.  Our model, RATA, is able to greatly outperform all other methods on both datasets.  Using the non-parametric knowledge of the table corpus is very advantageous for the large and specific vocabulary of entities in key columns. 

\begin{table}[!htb]
\centering
\begin{NiceTabular}{l|rrrr}
\hline
& \multicolumn{2}{c}{\textbf{EntiTables}} & \multicolumn{2}{c}{\textbf{WebTables}} \\
& MRR & NDCG & MRR & NDCG \\
\hline
TaBERT* & 56.0 & 46.4 & - & - \\
TABBIE* & 57.2 & 47.1 & - & - \\
TABBIE\dag & 25.18 & 15.2 & 12.44 & 11.93 \\
BART & 45.30 & 32.76 & 29.25 & 19.30 \\
RAG & 56.95 & 43.48 & 33.20 & 22.23 \\
RATA & \textbf{77.15} & \textbf{60.34} & \textbf{45.13} & \textbf{26.70} \\
& {\small $\pm$2.32} & {\small $\pm$2.18} & {\small $\pm$1.10} & {\small $\pm$0.73} \\
\hline
\end{NiceTabular}
\caption{Test results for row population, $n_{seed} = 2$. \\
* As reported in \citet{tabbie} \dag~Our results} 
\label{tab:row_pop}
\end{table}

\begin{table}[!htb]
\centering
\begin{NiceTabular}{l|rrrr}
\hline
& \multicolumn{2}{c}{\textbf{EntiTables}} & \multicolumn{2}{c}{\textbf{WebTables}} \\
& MRR & NDCG & MRR & NDCG \\
\hline
TaBERT* & 60.1 & 54.7  & -  & - \\
TABBIE* & 62.8 & 55.8 & -  & - \\
TABBIE\dag & 63.9 & 55.8 & 84.1 & 78.96 \\
BART & 73.36 & 65.37 & 87.40 & 85.05 \\
RAG & 78.64 & 72.81 & 89.39 & 87.58 \\
RATA & \textbf{88.12} & \textbf{81.01} & \textbf{94.07} & \textbf{89.94} \\
& {\small $\pm$1.91} & {\small $\pm$1.97} & {\small $\pm$0.44} & {\small $\pm$0.49} \\
\hline
\end{NiceTabular}
\caption{Test results for column population, $n_{seed} = 2$. \\
* As reported in \citet{tabbie} \dag Our results}
\label{tab:col_pop}
\end{table}

\begin{table}[!h]
\centering
\begin{NiceTabular}{l|rrrr}
\hline
& \multicolumn{2}{c}{\textbf{EntiTables}} & \multicolumn{2}{c}{\textbf{WebTables}} \\
& MRR & NDCG & MRR & NDCG \\
\hline
TABBIE & 10.62 & 11.56 & 24.79 & 26.17 \\
BART & 21.25 & 22.48 & \textbf{37.06} & \textbf{39.19} \\
TMatch & 30.54 & 32.23 & - & - \\
RAG & 18.65 & 19.71 & 34.80 & 36.34 \\
RATA & \textbf{34.32} & \textbf{36.25} & 33.58 & 35.33 \\ 
& {\small $\pm$2.80} & {\small $\pm$2.82} & {\small $\pm$1.60} & {\small $\pm$1.61} \\
\hline
\end{NiceTabular}
\caption{Test results for cell filling.}
\label{tab:cell_filling}
\end{table}

\begin{figure*}[th!]
  \centering
   \includegraphics[width=0.9\linewidth]{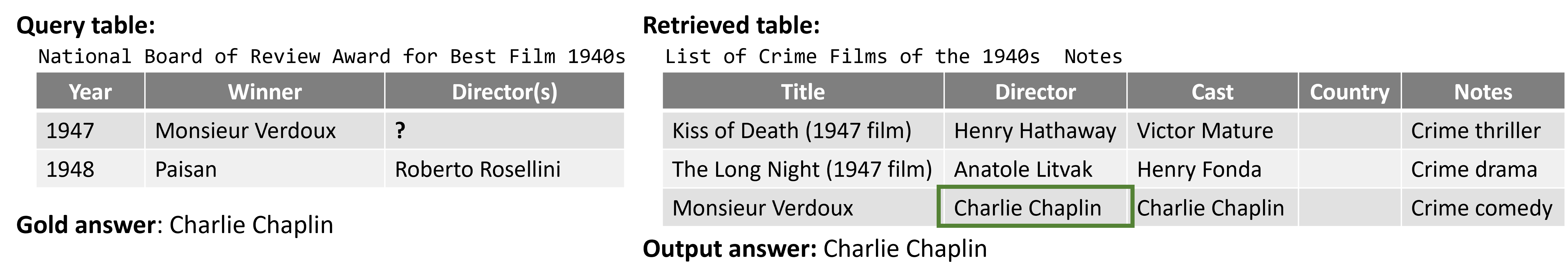}
   \caption{RATA example output on EntiTables dataset. The output answer is correct, and the retrieved table provides sufficient evidence for the answer.}
   \label{fig.case.cell_filling.entitables}
\end{figure*}

\begin{table*}[!ht]
\centering
\begin{tabular}{l|rr|rr|rr}
\hline
& \multicolumn{2}{c|}{\textbf{Row Population}} & \multicolumn{2}{c|}{\textbf{Column Population}} & \multicolumn{2}{c}{\textbf{Cell Filling}} \\
& EntiTables & WebTables & EntiTables & WebTables & EntiTables & WebTables \\
\hline
BM25 & 54.44{\small $\pm$2.72} & 41.16{\small $\pm$1.06} & 62.93{\small $\pm$2.73} & 84.17{\small $\pm$0.65} & 28.98{\small $\pm$2.59} & 38.48{\small $\pm$1.62} \\ 
DTR (initial) & 74.34{\small $\pm$2.39} & 47.88{\small $\pm$1.10} & \textbf{90.07}{\small $\pm$1.79} & \textbf{94.91}{\small $\pm$0.41} & \textbf{34.78}{\small $\pm$2.72} & \textbf{40.80}{\small $\pm$1.64} \\
DTR (post-RATA) & \textbf{80.98}{\small $\pm$2.17} & \textbf{49.62}{\small $\pm$1.11} & \textbf{90.97}{\small $\pm$1.72} & \textbf{94.94}{\small $\pm$0.41} & \textbf{37.48}{\small $\pm$2.81} & \textbf{40.26}{\small $\pm$1.66} \\
\hline
\end{tabular}
\caption{Retrieval answer-bearing MRR (\%).}
\label{tab:retrieval_all}
\end{table*}

Table \ref{tab:col_pop} contains our results for the column population task. RATA is again substantially better than the other methods, although not by as wide a margin as the row population task. The BART baseline is the best performing of the alternatives with an MRR lower by 6\% to 15\%.

Results on cell filling task are in Table \ref{tab:cell_filling}. Our method outperforms all baselines on both datasets. TABBIE performs the worst due to the large classification vocabulary and out-of-vocabulary issue. On EntiTables dataset, retrieval-based methods including TMatch and RATA significantly outperform non-retrieval methods including TABBIE and BART. Figure \ref{fig.case.cell_filling.entitables} shows an example output from RATA.
On WebTables, however, BART outperforms RATA. We notice that BART can achieve high scores by either copying values from other rows (as in Figure \ref{fig.webtable_case_1} and Figure \ref{fig.webtable_case_2}), or producing values similar with in other rows (as in Figure \ref{fig.webtable_case_3} and Figure \ref{fig.webtable_case_4}). As shown in the examples, this strategy is able to achieve good performance. 

\begin{figure}[tbh!]
  \centering
   \includegraphics[width=0.85\linewidth]{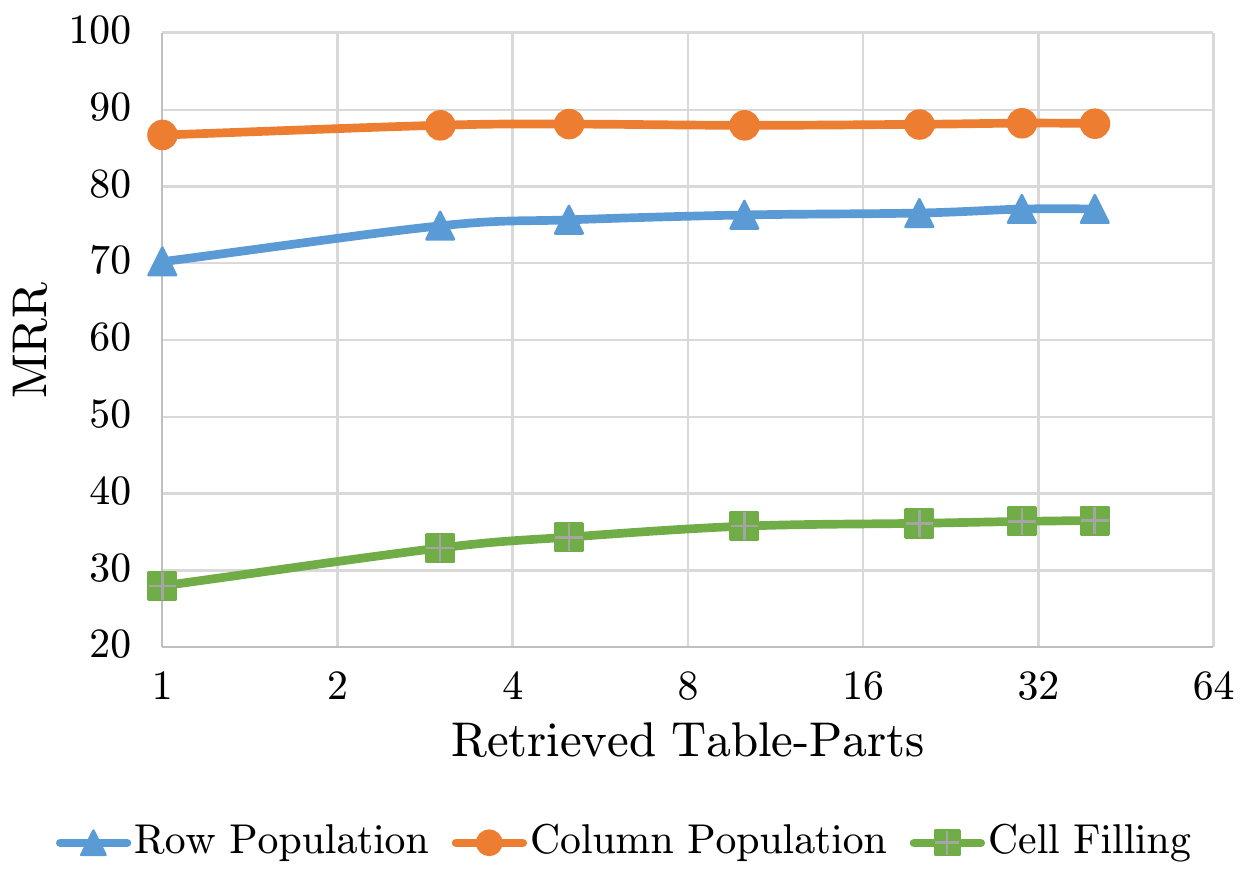}
   \caption{MRR gain as number of retrieved table-parts increases on the EntiTables dataset}
   \label{fig.ndoc}
\end{figure}

\paragraph{Effect of Retrieval}
To analyze the effectiveness of the DTR component, we report answer bearing MRR in Table \ref{tab:retrieval_all}. 
We notice that DTR is well trained after the initial retrieval training phase and achieves higher answer bearing MRR compared to BM25. End-to-end training provides meaningful supervision for retrieval and further improves MRR on most tasks.
By comparing Table \ref{tab:row_pop}, \ref{tab:col_pop}, \ref{tab:cell_filling} with Table \ref{tab:retrieval_all}, we notice that the final task MRR is close to answer bearing MRR. When the correct answer is present in the retrieved table, the reader can select the correct answer at high accuracy. This indicates that the bottleneck of our system is retrieval.

\paragraph{Number of Retrieved Table-Parts}
RATA was trained with 5 retrieved table-parts for all tasks. This relatively small number for the retrieval size provides good efficiency during training, since train time scales roughly linearly with the number of query / table-part pairs that must be processed by the reader transformer component.  But during inference, we are able to adjust the number of retrieved table-parts more freely. Figure \ref{fig.ndoc} shows that table augmentation performance monotonically increases as more evidence is retrieved for row population and cell filling, but column population performance does not improve past 5.

\FloatBarrier

\section{Conclusion}
Our retrieval-based transformer architecture for table augmentation, RATA, is able to greatly advance the state-of-the-art in three table augmentation tasks: row population, column population, and cell filling.
The non-parametric knowledge in the table corpus is able to substantially enhance the table augmentation capabilities.
Furthermore, by training an effective table-to-table retrieval model we are able to provide provenance for the system's proposed augmentations.
We also introduce a new benchmark dataset for table augmentation: \webtables{} and evaluate our model and two recent transformer baselines. Our code for RATA and the newly introduced dataset are available as open source\footnote{\url{https://github.com/IBM/retrieval-table-augmentation}}.

\section*{Limitations}


A limitation of RATA is always assuming the answer is included in the retrieval corpus, which is not always true. When the corpus does not contain the correct answer, the desired behavior is to inform the user that the answer cannot be obtained, but RATA will provide a poorly supported answer.
This also encourages RATA to learn spurious correlations when the retrieved tables coincidentally contain the same value, but does not really support the answer.
This problem is especially serious when the answer is very generic (for example, numbers like ``0'') and same values by coincidence are common.
This is related to the answerable question issue \citep{squad2} or evidentiality issue \citep{lee-etal-2021-robustifying,asai-etal-2022-evidentiality} for question answering.

\begin{figure}[tbh!]
  \centering
   \includegraphics[width=\linewidth]{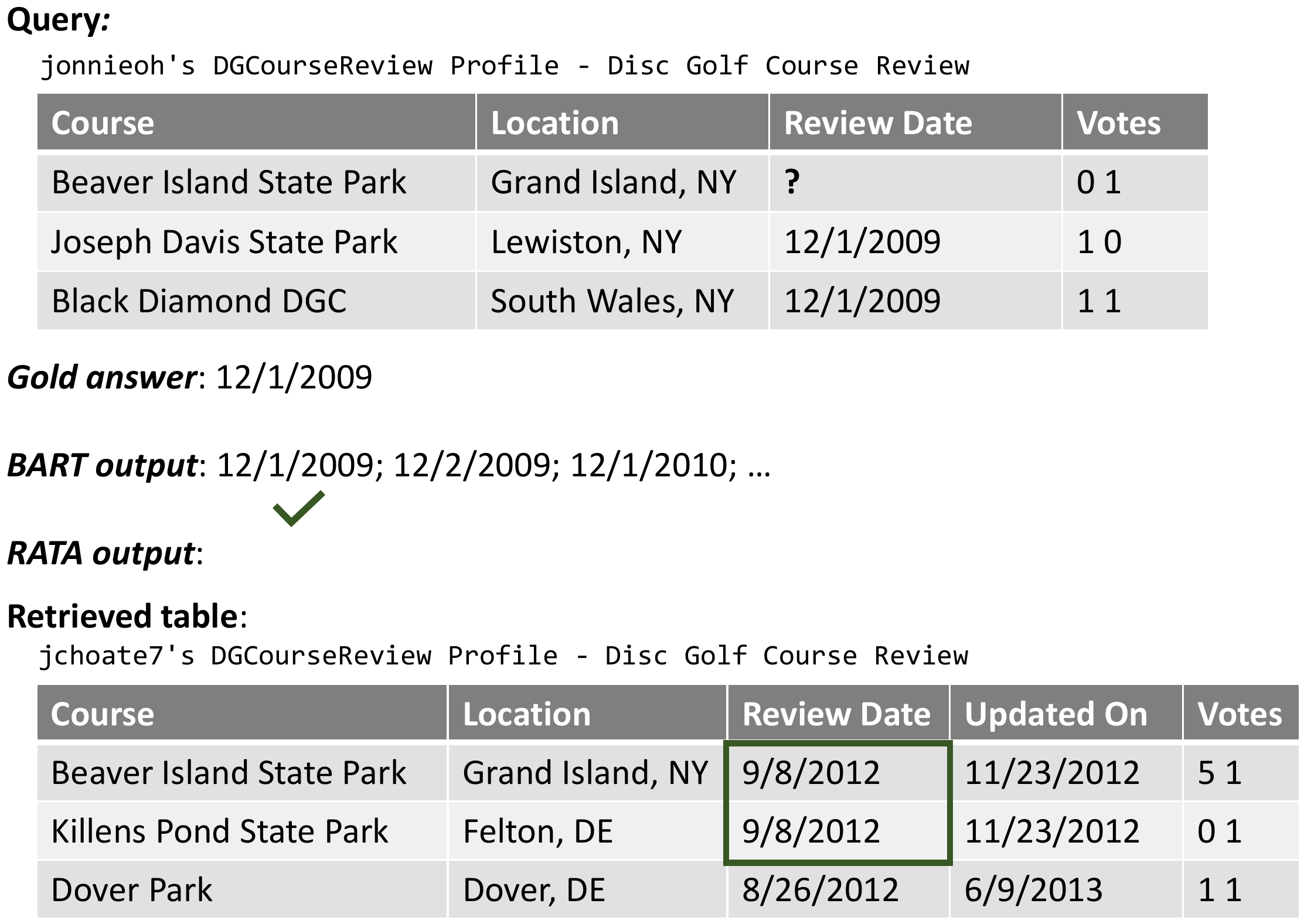}
   \caption{BART and RATA example outputs on WebTables.}
   \label{fig.webtable_case_1}
\end{figure}

For cell-filling on WebTables, BART outperforms RATA often by either copying values from other rows of the query table or producing values similar to those in other rows.
However, as shown in Figure \ref{fig.webtable_case_1}, RATA's retrieval is often not helpful. Usually, the information required to fill the query table is not repeated in the corpus, so the retrieved table cannot support the query. As a result, RATA is simply retrieving some similar table, and selecting similar values in the tables.

\bibliography{tblaug.bib}
\bibliographystyle{acl_natbib}

\appendix

\section*{Appendix}

\section{Model Hyperparameters}\label{apx.hypers}
Our model is fine-tuned from two BERT$_{\text{BASE}}$ models for the retriever and one BERT$_{\text{LARGE}}$ model for the reader. This totals $2 \cdot 110M + 340M = 560M$ parameters. 

Table \ref{tbl.hypers} shows the hyperparameters used in our experiments. 


\begin{table}[tbh!]
\begin{center}
\begin{tabular}{rrr}
\textbf{Hyperparameter} & \textbf{DTR} & \textbf{Reader}  \\
\hline
learn rate  & 5e-5 & 3e-5  \\
batch size & 128 & 32  \\
epochs & 3 & 2  \\
warmup instances & 0 & 10\% \\
learning schedule & linear & triangular \\
max grad norm & 1 & 1 \\
weight decay & 0 & 0 \\
Adam epsilon & 1e-8 & 1e-8
\end{tabular}
\end{center}
\caption{RATA hyperparameters}
\label{tbl.hypers}
\end{table}

The only hyperparameter that varied for the tasks and datasets was the batch size.
\begin{table}[tbh!]
\begin{center}
\begin{tabular}{rrr}
\textbf{Dataset} & \textbf{Task} & \textbf{Batch Size}  \\
\hline
Entitables & All & 32  \\
WebTables & Row Population & 32  \\
WebTables & Column Population & 32  \\
WebTables & Cell Filling & 64  \\
\end{tabular}
\end{center}
\caption{Batch size per task and dataset}
\label{tbl.hypers_batch_size}
\end{table}

\section{Dataset and Task Specifics}\label{apx.dataset_details}

We use two types of answer normalization. For EntiTables column population we implement case-insensitive matching by normalizing both predictions and gold answers to lowercase. For all row and column population in \webtables{} we use a normalization that removes unicode accents and non-ASCII characters then lowercases. Cell filling does not use normalization. 

For reproduction of results from TABBIE on Entitables we carry out the following steps.\\
\textbf{Column Header Population} Based on the above mentioned normalization we create a vocabulary of 182,909 column headers for the Entitables dataset which is approximately equal to the 127,656 possible header labels mentioned in the paper ~\citep{tabbie}. Each of the possible headers occurs atleast twice in the training dataset.\\
\textbf{Row Population} Except for above mentioned normalization we use entities which have occurred atleast 7 times in the training dataset which lead to 308,841 possible entities. THis is approximately equal to the 300,000 entities mentioned in ~\citep{tabbie}.  \\
\textbf{Cell Filling} Except for the above mentioned normalization we use cell values which have occurred atleast 10 times in the training dataset.\\

\section{Cell Filling BART Examples}

\begin{figure}[t!]
  \begin{subfigure}[b]{\linewidth}
    \centering
    \includegraphics[width=\linewidth]{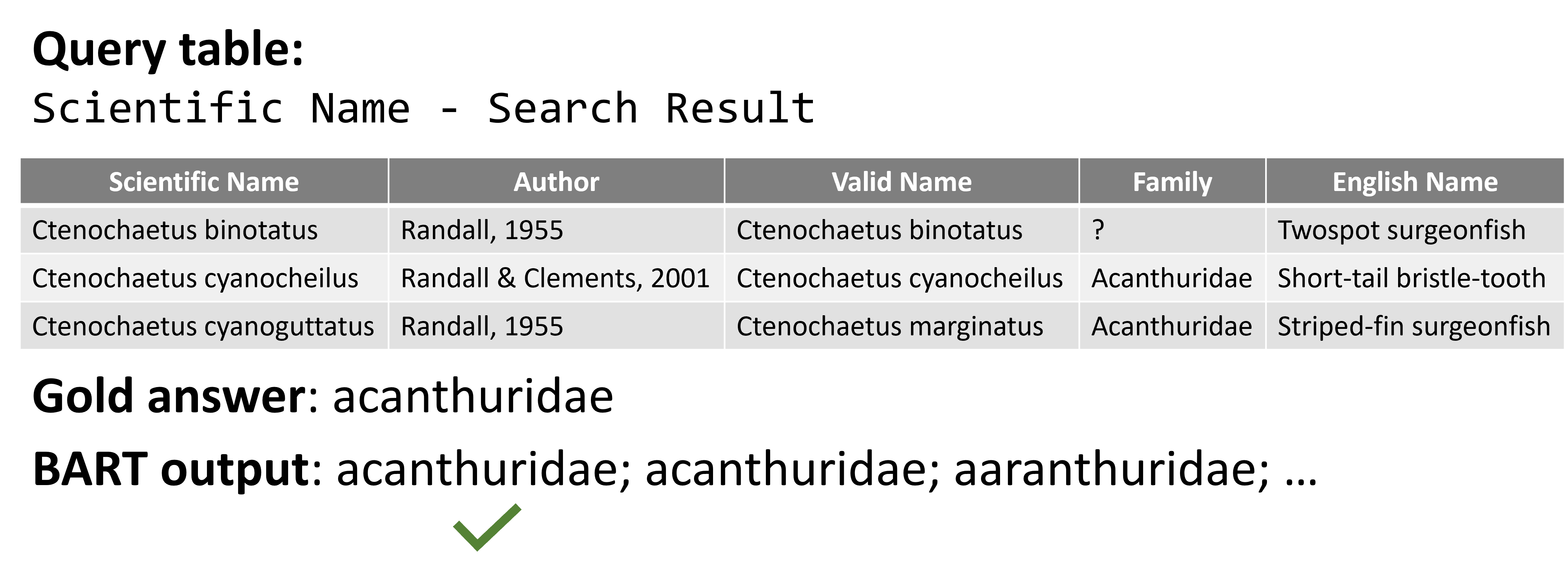}
    \caption{Additional example 1.}
    \label{fig.webtable_case_2}
  \end{subfigure}
    \begin{subfigure}[b]{\linewidth}
    \centering
    \includegraphics[width=\linewidth]{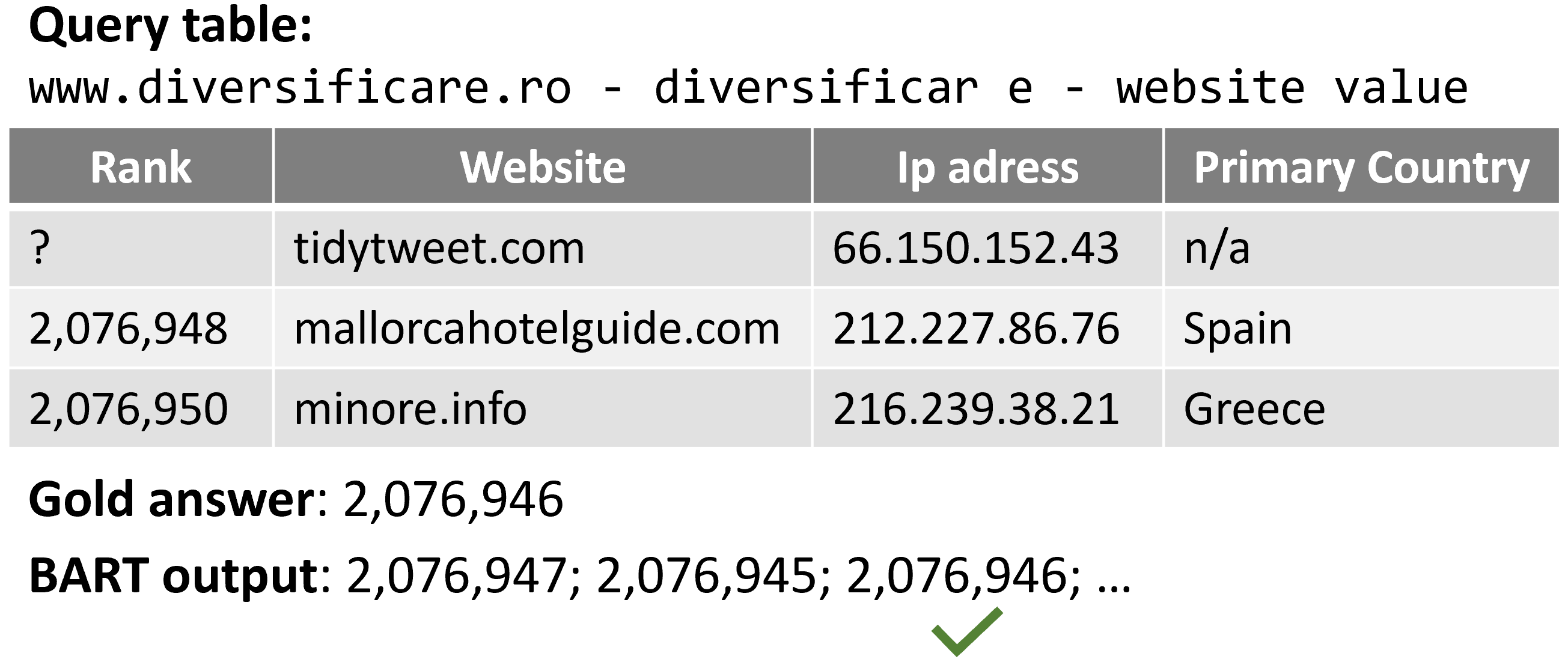}
    \caption{Additional example 2.}
    \label{fig.webtable_case_3}
  \end{subfigure}
  \begin{subfigure}[b]{\linewidth}
    \centering
    \includegraphics[width=\linewidth]{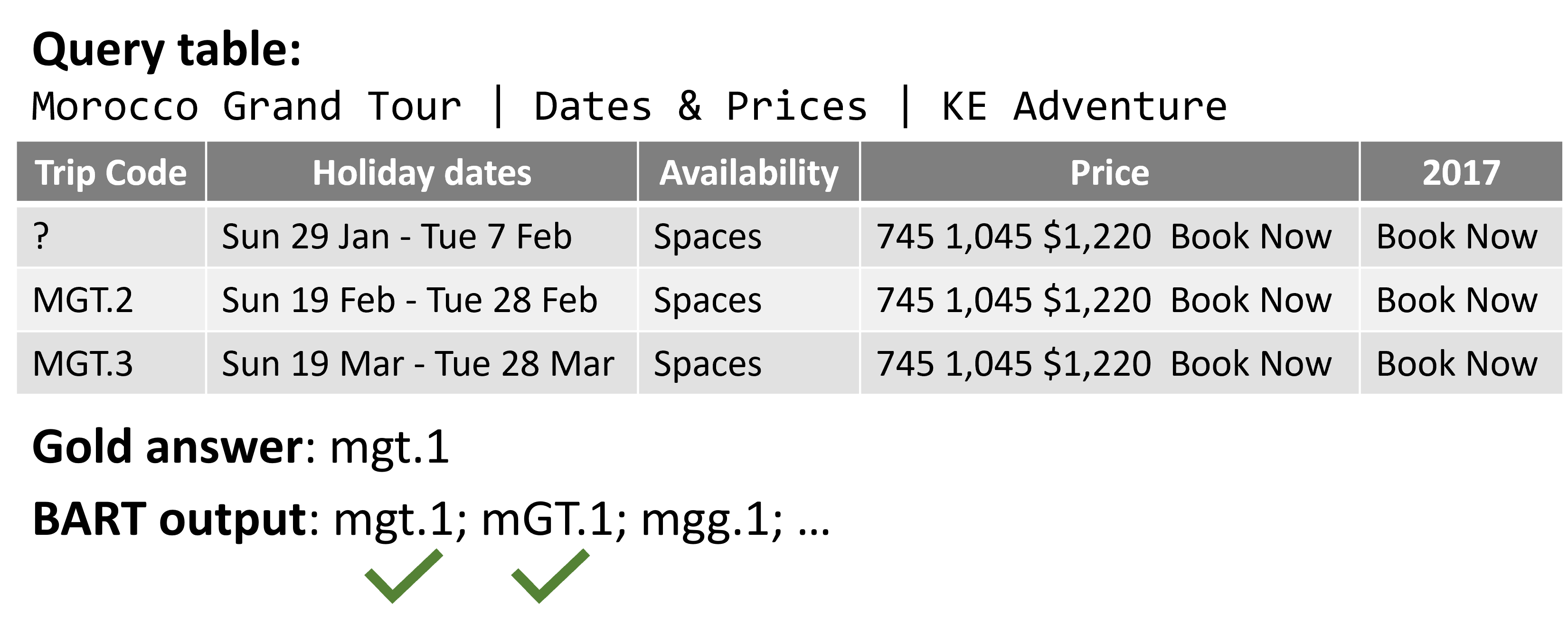}
    \caption{Additional example 3.}
    \label{fig.webtable_case_4}
  \end{subfigure}
  \caption{Additional BART output examples on WebTables dataset.}
  \label{fig.webtable_cases}
\end{figure}

Additional BART cell filling output examples on WebTables dataset are in Figure \ref{fig.webtable_cases}.

\section{Compute Infrastructure}

All row and column population experiments were done on a single P100 GPU. 
This gave train times of 24 to 48 hours. 
All cell filling experiments were done on a single A100 GPU, with train times of 24 hours. 

\end{document}